\documentclass[pdflatex,sn-chicago,round]{sn-jnl}

\jyear{2023}%
\usepackage[utf8]{inputenc}
\usepackage[T2A]{fontenc}
\usepackage{placeins}
\usepackage{array}
\usepackage{varwidth}
\usepackage{rotating}

\theoremstyle{thmstyleone}%
\theoremstyle{thmstyletwo}%

\theoremstyle{thmstylethree}%

\begin{document}

\title[A Survey of Resources and Methods for NLP of Serbian Language]{A Survey of Resources and Methods for  Natural Language Processing of Serbian Language}

\author[1]{\fnm{Ulfeta} \sur{Marovac}}\email{umarovac@np.ac.rs}

\author*[1]{\fnm{Aldina} \sur{Avdić}}\email{apljaskovic@np.ac.rs}

\author[2,3]{\fnm{Nikola} \sur{Milošević}}\email{nikola.milosevic@bayer.com}

\affil[1]{\orgdiv{Department of Technical and Technological Sciences}, \orgname{State University of Novi Pazar}, \orgaddress{\street{Vuka Karadžića 9}, \city{Novi Pazar}, \postcode{36300}, \country{Serbia}}}

\affil*[2]{\orgdiv{Research and Development}, \orgname{Bayer Pharmaceuticals}, \orgaddress{\street{Müllerstrasse 178}, \city{Berlin}, \postcode{13353},  \country{Germany}}}

\affil[3]{\orgname{ The Institute for Artificial Intelligence Research and Development of Serbia}, \orgaddress{\street{Fruškogorska 1}, \city{Novi Sad}, \postcode{21102},  \country{Serbia}}}



\abstract{The Serbian language is a Slavic language spoken by over 12 million speakers and well understood by over 15 million people. In the area of natural language processing, it can be considered a low-resourced language. Also, Serbian is considered a high-inflectional language. The combination of many word inflections and low availability of language resources makes natural language processing of Serbian challenging. Nevertheless, over the past three decades, there have been a number of initiatives to develop resources and methods for natural language processing of Serbian, ranging from developing a corpus of free text from books and the internet, annotated corpora for classification and named entity recognition tasks to various methods and models performing these tasks. In this paper, we review the initiatives, resources, methods, and their availability. }


\keywords{natural language processing, text mining, language resources, Serbian language}

\maketitle

\section{Introduction}\label{sec1}

The Serbian language is a south Slavic language currently actively spoken by about 12 million people worldwide. It is one of four mutually intelligible varieties of pluricentric language called Serbo-Croatian (other varieties include Croatian, Bosnian, and Montenegrin). Serbo-Croatian languages are morphologically rich \citep{delic2010speech}, containing many inflections of words, due to three genders, seven cases for nouns, and seven tenses for verbs, whose inflections are followed by other parts of speech and word types, as well as twelve sound changes occurring in word inflections  \citep{klajn2005gramatika}. Serbian is also the only European language that is formally digraphic and whose speakers are functionally digraphic, using both Cyrillic and Latin alphabets \citep{magner2001digraphia}. The majority of Serbian speakers live in Serbia (6,330,919 based on 2011 census\footnote{\url{https://data.stat.gov.rs/Home/Result/3102010401?languageCode=en-US}}), but a significant number of speakers also live in Montenegro, Bosnia and Herzegovina, Croatia, Macedonia, Slovenia, Albania, Hungary, Austria, Sweden, Germany, and other countries. The Serbian language is an official language in Serbia, Bosnia and Herzegovina, Montenegro, while it is recognized as a minority language in Croatia, Macedonia, Romania, Hungary, Slovakia, and the Czech Republic. Variants of the Serbo-Croatian language (Serbian, Croatian, Bosnian, and Montenegrin) are spoken by about 19 million people, and therefore the importance of these languages are quite significant \citep{ethnologue}. 

Natural language processing is a branch of artificial intelligence that examines methods to analyze, process and ultimately make natural languages understandable for computers \citep{reshamwala2013review}. Therefore, the field is addressing many challenges related to human/natural languages. Even though a majority of work in the field is predominantly done on the English language, there has been also work on other languages.  

High morphological complexity, variety of word inflection, and relatively low amount of resources available for Serbian and Serbo-Croatian pose a challenge for natural language processing and language technologies. The morphological richness of Serbo-Croatian makes it particularly interesting for examining how natural language processing methods perform on languages with a variety of inflection and how to efficiently handle word inflection in morphologically rich and low-resource languages. In sense of language technologies and natural language processing, Serbian cannot be viewed in isolation, as differences between Serbian, Croatian, Bosnian, and Montenegrin are small, and often approaches developed for one of these variants would perform well on others. Despite these challenges, there have been several initiatives, organizations, and significant academic work performed to address some of the specific challenges in Serbian. A number of resources and corpora for syntactic analysis, classification, and named entity recognition were developed, as well as a number of approaches for document analysis, classification, semantic similarity, and even analysis of rhetorical figures such as similes. 

The development of digital lexical resources is an important and strategic task for every language and should have national priority. The results of natural language processing are dependent on the quality and volume of available digital resources, as well as the availability and comprehensiveness of tools for processing digital resources \citep{nenadic2004creating}. Our goal is to collect available resources and methods for processing textual data in the Serbian language, describe them, and identify shortcomings that can be advanced and expanded with the most needed resources according to the development trends of NLP. To the best of our knowledge, this is the first review of NLP resources and methods for Serbian of newer date and scope. 

\subsection*{A brief history of NLP resources and method development in Serbia}

The development of the first digital corpora in the Balkans started shortly after the development of the first digital corpora in the world, and it was started by the psychologist Djordje Kostić, in 1957. with the goal of developing language technologies for speech recognition and machine translation from the Serbo-Croatian language. This corpus was developed until 1962, however, it was not digitally processed, so the first digital corpus was published in Zagreb in 1967. This corpus contained the epic Osman by Ivan Gundulić prepared by Željko Bujas. Development of corpora and corpus linguistics in the western Balkans in the period between 1950 and 1990 is presented by \cite{dobric2012savremeni}. Language resources and tools that were mainly developed at the Faculty of Mathematics, University of Belgrade, until the year 2003, have been previously reviewed \citep{vitas2003processing, vitas2003overview}. During the project called META-NET in 2012, the analysis of the language resources for 23 official languages of the European Union was done, and as a part of white pages was published book "The Serbian language in the digital age" \citep{duvsko2012srpski}. The Regional Linguistic Data Initiative (ReLDI) project has made a significant contribution to promoting the relevance and importance of open language resources for Serbian and related languages \citep{samardvzic2015regional}. The open and freely available language resources for processing the Serbian language, developed within the ReLDI project or independently built, are briefly presented by \cite{batanovic2020otvoreni}.

In this paper, we aim to review corpora, resources, methods, models, and tools that were developed over time for the Serbian language. We intentionally limit this review to the Serbian language only. While we agree that some approaches do work as well on related languages in the Serbo-Croatian group of languages, there are still small differences between them, that would make evaluation and comparison of the resources and methods challenging.

\section{Review methodology}

To cover all authors who deal with natural language processing in Serbian, we started with the National Repository of Dissertations in Serbia (\url{https://nardus.mpn.gov.rs/}). By using keywords such as natural language processing, NLP, text mining, text data processing, computational linguistics, electronic dictionaries, corpora, sentiment analysis, emotional analysis, text classification, lexical resources, and other synonyms and related terms, we identified dissertations that contain these keywords. From the most relevant dissertations, those that deal with natural language processing in Serbian were selected. Additionally, a set of dissertations were identified by searching for known NLP scientists, supervisors, and groups at Serbian universities. Dissertations were identified by searching for known scientists acting as a supervisor or a member of the thesis committee. A total of 29 dissertations in the field of natural language processing were selected. By analyzing the dissertations and references cited in them, we identified 316 papers indicating NLP for Serbian. Further searches were conducted on Google Scholar for prominent authors (or author groups) and selected topics.  

We reviewed the dissertations and papers we identified, excluding those that did not pertain to natural language processing in Serbian, and classified them based on their topic and date of publication. In this review, we follow this classification, with each section covering a broad area of natural language processing. The content in each section is primarily arranged in chronological order.

\section{Corpora}
A corpus is a set of machine-readable texts representing a sample of a language or text type.
Corpora can be classified based on their parameters such as medium, scope (size), domain, purpose, period, source, method of annotation, number of languages involved, etc. \citep{vitas2003processing}. Given that corpora can include texts in one or more languages, they are divided into monolingual and multilingual corpora. According to this classification, we will present the corpora of the Serbian language. 

\subsection{ Monolingual corpora}
\textit{The Diachronic Corpus of the Serbo-Croatian Language} (DCSCL, Table \ref{tabcorp}) of Professor Kostić digitized in 2003 contains texts from the period from the 12th to the 20th century, divided into five-time samples. The corpus comprises 11 million words that have been manually annotated with lemmas and information on various morphological categories such as gender, number, case, person, tense, and more \citep{akostic}.
In 1981, the NLP group at the Faculty of Mathematics (NLP\_MATF) initiated the development of a corpus for the contemporary Serbian language. The first version of this corpus, named \textit{"The Untagged Corpus of Contemporary Serbian Language"} (UCCSL, Table \ref{tabcorp}), was created in 2003. This corpus contains literature published during or after the 20th century and lacks any annotations. Subsequently, the corpus was enhanced by incorporating bibliographic information into the corpus texts, and this new version was called \textit{"SrpKor2003"} (SrpKor2003, Table \ref{tabcorp}) \citep{krstev2005corpus, utvic2014izgradnja}.

Most of the monolingual corpora have morphosyntactic annotation and bibliographic information about the corpus texts. Morphosyntactic annotation is a linguistic annotation that adds tags to the token: type of speech (Part of Speech Tagging), canonical form or lemma (lemmatization), and morphological word categories. 
By expanding SrpKor2003, a new version of the corpus of contemporary Serbian \textit{"SrpKor2013"} (SrpKor2013, Table \ref{tabcorp}) was created, which contains literary and artistic texts by Serbian writers in the 20th and 21st centuries, as well as scientific texts, administrative texts, and general texts. \textit{The Corpus of Contemporary Serbian} contains bibliographic data and it has been automatically morphosyntactically annotated (with part-of-speech and lemmas). It contains more than 122 million words. Its subset \textit{"Lematized Corpus of the Modern Serbian Language"} (SrpLemKor, Table \ref{tabcorp}) contains 3.7 million corpus words. Both corpora are available with registration under a license \citep{popovie2010taggers, utvic2011annotating}.

\begin{sidewaystable}[!htbp]
\tiny
\centering
\caption{Monolingual Serbian corpora }\label{tabcorp}%
\begin{tabular}{@{}p{0.18\linewidth}p{0.2\linewidth} lll@{}}

\toprule
Corpus label & Text type & Number of unit & Annotation \footnotemark[1]  & Reference  \\
DCSCL   & general   & 11 000 000 words & S, L, M;MA  & \text{\citep{akostic}} \\
  UCCSL & literary   & 22 200 000 words & U  & \text{\citep{krstev2005corpus, utvic2014izgradnja}} \\
  SrpKor2003 & literary   & 22 200 000 words & B; MA &\text{\citep{krstev2005corpus, utvic2014izgradnja}} \\

 SrpKor2013 & literary-artistic
scientific   & 122 000 000 words & B,L,PoS; MA,AA  & \text{\citep{popovie2010taggers, utvic2011annotating}} \\
SrpLemKor & literary-artistic
scientific   & 3 700 000 words & B,L,PoS; MA,AA  & \text{\citep{popovie2010taggers, utvic2011annotating}} \\
HennC & literary  & 728 952 words & B; MA  & \text{\citep{Henning1700}} \\
UnVukC
 & literary   &   6919 proverbs & U & \text{\citep{krstev1997jedan}} \\
 VukSimC
 & literary   &   852 similes  & U & \text{\citep{milosevic2016cool,milosevic2018creating}} \\
 MRCOR1
 & medical reports   &  2212 reports & MT,NMT; MA  & \text{\citep{avdic2021realizacija}} \\
 MRCOR2
 & medical reports   &  2000 reports & MT,NMT; MA  & \text{\citep{avdic2021realizacija}} \\
  
DMRC
 & medical reports   &  150 reports & U  & \text{\citep{jacimovic2015rule}} \\
 LAWC
 & text of laws   &  59167 texts & S; MA  & \text{\citep{petrovic2019influence}} \\
 LTC
 & text of laws   &  7981446 tokens & S,NE; MA  & \text{\citep{vasiljevic2015}} \\

ATC
 & newspapers
scientific   &  200 000 words & M; MA  & \citep{secujski2008software} \\
SrpNEval
 & news  &  89425 words & NE; AA, MA  & \text{\citep{krstev2012recognition}} \\

NormTagNER
 & social network   &   89 425 words & M,NE;MA, AA & \text{\citep{milivcevic2016tviterasi}} \\
 SETimes.SR
 & news   &   86 726 tokens & S, L, PoS, SD, NE; MA & \text{\citep{batanovic2018setimes, batanovic2018training}} \\
 paraphrase.sr& news   & 1194  pairs  & SS; MA & \text{\citep{Batanovic2011softverski}} \\
 STS.news.sr & news   & 1192 pairs  & SS; MA & \text{\citep{batanovic2018fine}} \\
SrELTeC & novel   &   5 263 071 words & S, L, PoS, NE; AA & \text{\citep{stankovic2021annotation}} \\
SrpELTeC-gold & literary   &   330 119 tokens &  NE; AA,MA & \text{\citep{todorovic2021serbian}} \\
SrpKor4Tagging & literary   &   342 803  tokens &  PoS, L; AA & \text{\citep{stankovic-etal-2020-machine}} \\
RudKorP & academic text   &   2 340 000 words &  PoS, L; AA & \text{\citep{utvic2019pretraga}} \\
TorlakKor & culture interview transcript  &   498021 tokens &  MS,A,L; MA,AA & \text{\citep{11356/1281}} \\
COPA-SR & question answer  &   1000 premises & P; MA & \text{\citep{11356/1708}} \\
CorFoA & biographical interviews transcripts  &   171552 tokens & MS, L; AA & \text{\citep{11356/1422}} \\
MLNews & news   &  659084 tokens & MS, L; AA & \text{\citep{11356/1371}} \\

MLN-COM & news comments  &  878482 tokens & MS, L; AA & \text{\citep{11356/1372}} \\

srWaC & web text  &  554627647 tokens & MS, L, SD; AA & \text{\citep{11356/1063, ljubevsic2016serbian}} \\

CorLeg & laws &  5 files & U & \text{\citep{11356/1754}} \\
\botrule
\end{tabular}
\footnotetext[1]{Annotation target: U - unannotated, MS - morphosyntactic, L - lemmatization, M - morphological categories, PoS - Part of Speech Tagging, S - structural, NE - named entity, UK - unknown annotation, A - accentuation, B - bibliographic, SD - syntactic dependencies, P - plausibility, MT - medical terms, NMT - non-medical terms, SS - semantic similarity;
Annotation type: AA - automated annotation MA - manually annotated }

\end{sidewaystable}

Among the available corpora of the Serbian/Serbo-Croatian language
at the Faculty of Mathematics of the University of Belgrade\footnote{\url{http://www.korpus.matf.bg.ac.rs/prezentacija/korpusi.html}}, there are also the following monolingual corpora. \textit{Henning's Corpus of Serbo-Croatian} (HennC, Table \ref{tabcorp}) consists of approximately 700,000 words of Serbo-Croatian. The texts are taken from modern
Yugoslav fiction and all Serbo-Croatian-speaking areas are represented (Serbia, Croatia, Montenegro, and Bosnia-Herzegovina) \citep{Henning1700}. \textit{The Untagged Corpus of Vuk's Folk Proverbs} (UnVukC, Table \ref{tabcorp}), containing folk proverbs along with Vuk's comments on them \citep{krstev1997jedan}. Besides this corpus, Vuk's collection of similes has been augmented by employing grammatical rules, machine learning, and manual review. As a result, a corpus of contemporary similes containing 852 similes was developed (VukSimC, Table \ref{tabcorp})\footnote{ \url{https://ezbirka.starisloveni.com}} \citep{milosevic2016cool,milosevic2018creating}. 
\textit{Electoral Crisis 2000} corpus, which includes the entire webcasts of the daily newspaper "Politika" from September 10th to October 20th, 2000, and the \textit{Labeled corpus of the Serbian language}, which consists of texts with a minimal set of structural labels, lack the detailed information on size and are available on the same source\footnote{\url{http://www.korpus.matf.bg.ac.rs/prezentacija/korpusi.html}}.

There are smaller corpora that have been collected mostly for specific
domains (medicine, law, etc.) and particular purposes (name entities recognition,  semantic similarity, etc.). Among them are the corpora \textit{MRCOR1} and \textit{MRCOR2} (Table \ref{tabcorp}) consisting of medical reports reviewed from 32
medical centers in Serbia. The primary data set contains 2212 medical reports with a diagnosis of measles. The other dataset consists of 2000 medical reports with ten different types of diagnoses. Medical and non-medical terms are manually marked in the medical reports. For each medical report is assigned a diagnosis code \citep{avdic2020automated}.
A corpus (DMRC, Table \ref{tabcorp}) of 100 discharge lists and 50 reports from doctors from the Faculty of Dentistry at the University of Belgrade was used to evaluate the system's effectiveness in automatically analyzing temporal expressions of medical narrative texts.  Previously, the texts had been automatically de-identified, but the time expressions had not been changed \citep{jacimovic2015rule}. The \textit{LAWC} (Table \ref{tabcorp}) set of data includes a collection of 1120 texts of laws, segmented into a total of 59167 texts of individual articles of law \citep{petrovic2019influence}. The corpus \textit{LTC} (Table \ref{tabcorp}) consists of legal texts in electronic form that are available on the website of the National Assembly of the Republic of Serbia. The laws passed by the end of May 2014 contain 681 texts \citep{vasiljevic2015}.

AlfaNum which deals with automatic speech recognition (ASR), has built its resource \textit{AlfaNum Text Corpus} (ATC, Table \ref{tabcorp}), characterized by morphological categories and accentuation and contains approximately 200,000 words \citep{secujski2008software}. 
\textit{The Named Entities Evaluation Corpus for Serbian} (SrpNEval, Table \ref{tabcorp}) consists of 2000 short news Serbian daily newspapers from 2005 and 2006. Both the Cyrillic and Latin official scripts for the Serbian language are used in the corpus \citep{krstev2012recognition}. 
\textit{ReLDI-NormTagNER-sr 2.1} (NormTagNER, Table \ref{tabcorp}) is a manually annotated corpus of Serbian tweets for evaluation of tokenization, sentence segmentation, word normalization, morphosyntactic tagging, lemmatization, and named entities recognition of non-standard Serbian language \citep{milivcevic2016tviterasi}. 
\textit{SETimes.SR} (SETimes.SR, Table \ref{tabcorp})  is a reference training corpus of Serbian, which has been annotated on multiple levels.  The texts in SETimes.SR were obtained from the multilingual parallel corpus \textit{SETimes} (SETimes, Table \ref{tabparcorp}), which is a collection of news articles from the now-defunct Southeast European Times news portal \citep{batanovic2018setimes}, \citep{ batanovic2018training}. 
Sentences from online press sources were collected for  \textit{The Serbian Corpus of Paraphrases} (paraphrase.sr, Table \ref{tabparcorp}). A binary similarity score was manually assigned to each pair of sentences, indicating whether the sentences in the pair are semantically similar enough to be considered close paraphrases \citep{Batanovic2011softverski}.
Another corpus for determining semantic similarity, \textit{The Serbian Corpus of Short News Texts} - (STS.news.sr, Table \ref{tabparcorp}), consists of 1192 pairs of sentences in Serbian that were collected from news sources on the internet \citep{batanovic2018fine}.

Old Serbian novels from the 1840s to the 1920s are collected in \textit{SrELTeC} (SrELTeC, Table \ref{tabcorp}) and have been digitally preserved as part of the COST action CA16204 \citep{stankovic2021annotation}. \textit{ELTeC}'s section for Serbian contains 120 novels \citep{odebrecht2021european}. The novels have structural annotations, and sentence splitting, words are POS-tagged, lemmatized and seven classes of named entities are annotated. Some of the other resources available through the ELG\footnote{\url{https://live.european-language-grid.eu/}}  portal are \textit{SrpELTeC-gold}, \textit{SrpKor4Tagging}, and \textit{RudKorP}  (Table \ref{tabcorp}) .
The corpus for training the recognition of named entities SrpELTeC-gold is a sub-corpus of the literary corpus of the Serbian language, marked with named entities by the SrpNER\citep{krstev2014system} system \citep{todorovic2021serbian}.
The SrpKor4Tagging corpus was formed by combining literary and administrative texts in the Serbian language \citep{stankovic-etal-2020-machine}.
 The RudKorP corpus contains texts in the field of mining and processing of mineral raw materials, created at the University of Belgrade, Faculty of Mining and Geology \citep{utvic2019pretraga}.
There are several more corpora available through the Clarin.si \footnote{https://www.clarin.si} platform, which are shown at the bottom of Table \ref{tabcorp}.
\textit{TorlakKor} is a corpus of transcripts of interviews with the local population of Timok (an area in southeastern Serbia) \citep{11356/1281}. The \textit{COPA-SR} dataset (\textit{Choice of Plausible Alternatives in Serbian}) is a translation of the English \textit{COPA} dataset \citep{11356/1708}. \textit{CorFoA} is a corpus of Serbian forms of the address containing transcripts of biographical interviews with 19 participants \citep{11356/1422}.
\textit{MLNews} is a comprehensive corpus of  news articles that are Serbian language-related. It is complemented with a separate corpus of citizens' online comments on the news articles, available as \textit{MLN-COM} \citep{11356/1371, 11356/1372}. The web corpus of the Serbian language \textit{srWaC} was built by crawling the .rs top-level domain for Serbia in 2014 \citep{11356/1063, ljubevsic2016serbian}. 
\textit{CorLeg} is a corpus of legislation texts of the Republic of Serbia which was created using a large number of Serbian Legislation texts gathered from the official website \footnote{ https://www.pravno-informacioni-sistem.rs/} \citep{11356/1754}.

\subsection{Multilingual corpora}

Multilingual corpora are a particular type of corpus that contains texts written in multiple languages. Parallel corpora include both the original texts and their translations into one or more other languages presented in such a way that their logical structure is explicitly connected at the document, chapter, paragraph, sentence, or word level. Table \ref{tabparcorp} shows multilingual corpora containing original texts or translations in the Serbian language.
One of the early attempts to develop multilingual corpora is the creation of an alignment corpus of Plato's \textit{"Republic"} containing translations into 21 languages, including Serbian. The corpus has been annotated at the sentence level and has been utilized for both tool development and automated alignment \citep{krstev2011aligned}. The multilingual language resources and tools for extracting information from the language corpora of CEE languages (Central and Eastern European Languages), called MULTEXT-East\footnote{http://nl.ijs.si/ME/} were created as part of the project Multext. The book "1984" is included in this parallel and sentence-aligned corpora, \textit{Multext-East Corpora} (G. Orwell's "1984")(G.O.1984, Table \ref{tabparcorp}), along with translations into several other languages. \cite{krstev2011aligned} created a translation of this novel into Serbian and a morphosyntactic annotation in the MULTEXT-East format, for which they had previously developed a specification for the Serbian language. The parallel corpus \textit{Verne80days} (Table \ref{tabparcorp}) contains the French original and 17 translations of Jules Verne's novel "Around the World in 80 Days". The alignment was performed on the sub-sentence level for each language \citep{vitas2008tour}. \textit{The Serbian-French Corpus} (SrpFranKor, Table \ref{tabparcorp}), which consists of 31 subsentence-aligned texts that were originally written in French and then translated into Serbian and vice versa, is the first bilingual corpus in the Serbian language \citep{vitas2006literature, krstevpreparation}. \textit{ParCoLab} (Table \ref{tabparcorp}) is a parallel online searchable corpus consisting of sentence-alignment texts in French, Serbian, English, Spanish, and Occitan. Each of these languages is at the same time a source language and a translation language \citep{balvet2014talc1}.
  
\begin{sidewaystable}[!htbp]
\tiny
\centering
\caption{Parallel Serbian corpora }\label{tabparcorp}%
\begin{tabular}{@{}p{0.10\linewidth}p{0.15\linewidth}p{0.10\linewidth}p{0.10\linewidth}p{0.13\linewidth}l@{}}
\toprule
Corpus label & Text type & Number of unit & Annotation\footnotemark[1]   & Languages & Reference  \\
Plato's Republic & philosophical text   & 21 text translation  & S, A; MA, AA  & multilingual & \text{\citep{vitas98plato}} \\
 G.O.1984 & general   & 100000 words & L, M, PoS, A;AA, MA  & multilingual & \text{\citep{krstev2011aligned}} \\
 Verne80days &   literary   & 32 aligned texts & L, M, PoS, A; AA,MA  & multilingual & \text{\citep{vitas2008tour}} \\
 SrpFranKor  & literary   &  1738752 words & S, A; AA,MA  & Serbian, French & \text{\citep{vitas2006literature, krstevpreparation}} \\
 ParCoLab  & general   &  32 000 000 words & S, A; AA,MA  & multilingual & \text{\citep{balvet2014talc1}} \\

SrpEngKo & general   & 4.420.711 words & S,A; AA,MA  & Serbian, English & \text{\citep{krstev2011aligned}} \\

SETimes & news   & 86 726 tokens & L,M,PoS; MA  & multilingual & \text{\citep{batanovic2018setimes}} \\
SELFEH & law, finance, education, and health
   & 2 000 000 words & L,M,PoS, A;AA, MA  & Serbian, 
 English & \text{\citep{utvic2011annotating}} \\
srenWaC 1.0 & web text 
   & 23139804 words &  A;AA & Serbian, English & \text{\citep{11356/1059}} \\
SrpNemKor & literary   & 1 657 329 words & S,A; AA,MA  & Serbian, German & \text{\citep{andonovski2019bilingual, andonovski2020mreza}} \\
OpenSubtitles & film translations   & 2 793 243 tokens & S,A; AA,MA  & multilingual & \text{\citep{Tiedemann2012parallel}} \\

BsHrSrWaC & web text   & 894  000 000 tokens & M,L,S,D; AA,MA  & multilingual & \text{\citep{ljubesicklubicka2014}} \\
Twitter-HBS & social network   & 390268 texts & S; AA,MA  & multilingual & \text{\citep{11356/1482}} \\
PE2rr & general   & 43938 words & S,ER,A; AA,MA  & multilingual & \text{\citep{11356/1065}} \\
BERTić-data & general   & 8387681518 words & S; MA  & multilingual & \text{\citep{ljubevsic2021berti}} \\
CLASSLA-Wiki & general   & 486258862 tokens & LN; AA  & multilingual & \text{\citep{11356/1427}} \\
\botrule
\end{tabular}
\footnotetext[1]{Annotation target: U - unannotated, L - lema, MS - morphosyntactic, M - morphological categories, LN - linguistic, PoS - Part of Speech Tagging, UK - unknown annotation, S - structural, SD - syntactic dependencies, A - aligned, ER - error;
Annotation type: AA - automated annotation MA - manually annotated }

\end{sidewaystable}

\textit{The Serbian-English Corpus} (SrpEngKo, Table \ref{tabparcorp})  is the second bilingual collection.  It consists of English source texts aligned with their translations into Serbian, and visa-versa, as well as several aligned English and Serbian translations of literary texts originally written in French \citep{krstev2011aligned}.
The corpus \textit{SETimes} is based on the articles posted on the news website SETimes.com. Bulgarian, Bosnian, Greek, English, Croatian, Macedonian, Romanian, Albanian, and Serbian are among the ten languages in which the news is available. Part of the SETimes, sub-corpus \textit{BALKANTIMES} was used for the expansion of SrpEngKo \citep{batanovic2018setimes}. Parallel texts from the fields of law, business, education, and health care are also added to SrpEngKor, resulting in the creation of the sub-corpus \textit{Serbian-English Law Finance Education and Health} (SELFEH, Table \ref{tabparcorp}). Almost 150 parallel texts make up SELFEH, which was utilized in term extraction and machine translation research as well as to test various taggers for the Serbian language \citep{utvic2011annotating}.
Another Serbian-English corpus is  \textit{srenWaC} (Table \ref{tabparcorp}), which consists of sentence-aligned parallel texts pulled from the .rs top-level domain \citep{11356/1059}.
In addition to the SrpFranKo and SrpEngKo bilingual corpora, a similar corpus was created for the German (SrpNemKor, Table \ref{tabparcorp}). It contains 48,004 translated pairs of literary texts in Serbian and German, which are aligned to the sentence level. Available tools for annotation of named entities in texts in both languages as well as tools for terminology extraction were applied to the prepared parallel corpus \citep{andonovski2019bilingual, andonovski2020mreza}.

Additionally, there are multilingual parallel corpora, some of which are displayed at the table's end  (Table \ref{tabparcorp}).
\textit{OpenSubtitles} is a database with about 4 million sentence-level translations of movies and television shows in more than 62 different languages \citep{Tiedemann2012parallel}. 
\textit{The Bosnian, Croatian, and Serbian Web Corpora} (BsHrSrWaC, Table \ref{tabparcorp}) are top-level-domain web corpora. They were used to create a method for separating similar languages that is based on unigram language modeling on the crawled data only \citep{ljubesicklubicka2014}.
The Twitter user dataset (Twitter-HBS, Table \ref{tabparcorp}) consists of tweets and their language tag (Bosnian, Croatian, Montenegrin, or Serbian). The main goal of creating this corpus is discrimination between closely related languages at the level of Twitter users \citep{11356/1482}.
The \textit{PE2rr} corpus includes source language texts from many fields, as well as automatically produced translations into a number of morphologically rich languages, post-edited versions of those texts, and error annotations of the post-edit processes that were carried out. This corpus contains texts in Spanish, German, Serbian, Slovene and English \citep{11356/1065}.
The  \textit{BERTić-data} text collection contains more than 8 billion tokens of mostly web-crawled text written in Bosnian, Croatian, Montenegrin, or Serbian. The collection was used to train the BERTić transformer model \citep{ljubevsic2021berti}.
The Wikipedia dumps of the Bosnian, Croatian, Macedonian, Montenegrin, Serbian, Serbo-Croatian, and Slovenian Wikipedias were collected in the comparable corpus  \textit{CLASSLA-Wikipedia} (CLASSLA-Wiki, Table \ref{tabparcorp}). The linguistic annotation was performed with the classla package \footnote{https://pypi.org/project/classla/}, on all levels available for a specific language \citep{11356/1427}.
Corpora for sentiment analysis are presented in a separate chapter.

\section{Language resources}

\subsection{Dictionaries and terminologies}

The term electronic dictionary considers the dictionary which is used for text processing.  It consists of valuable information for solving problems of segmentation, morphological, and partly syntactic and  semantic text processing \citep{vitassrpski}. The automatic processing of text  begins by analyzing individual words, which are the base units of the analyzed text. At times, individual words may not be the most appropriate base units for processing natural language. Therefore, there are two types of dictionaries: mono-lexemic, which consists of single words,  and polylexemic which consists of multi-word units \citep{andonovski2020mreza}.

The international network of laboratories for computational linguistics, RELEX \citep{laporte2003relex}, has created a model for building electronic morphological dictionaries that have been adopted by numerous organizations dealing with natural language processing. The \textit{Unitex} \footnote{https://unitexgramlab.org/language-resources} system works with electronic morphological dictionaries developed according to this model. These are dictionaries in DELA format (\textit{Dictionnaires Electroniques du LADL - Laboratoire d’Automatique Documentaire et Linguistique}). In order to distinguish between monolexemic and polylexemic units, this electronic dictionary is organized into two separate subsystems: a dictionary of monolexemic units (DELAS - simple forms and DELAF - inflected forms) and a dictionary of polylexemic units (DELAC - compound forms, and DELACF - compound inflected forms). 

Based on these models, within the Group for Language Technologies of the University of Belgrade, electronic morphological dictionaries of the Serbian language in Latin and Cyrillic (SrbMD) were built \citep{krstev1997jedan, vitas2003processing, vitas2003overview, krstev2006prerequisites, krstev2010automatic}. 
According to \citep{mladenovic2016}, the SrpMD system currently contains 148,000 lemmas and over 1,000 final transducers that generate more than 5 million DELAF determinations. The tool Leximir \citep{stankovic2011production} is used as a dictionary management system. It is
a multipurpose tool for supporting computational linguists in developing, maintaining, and exploiting e-dictionaries. 

The accentuation-morphological dictionary was created at the Faculty of Technical Sciences in Novi Sad and it contains over 4 million entries. It is used for context analysis within text-to-speech and automatic speech recognition systems for Serbian \citep{secujski2008software}.

 \cite{ljubevsic2015mwelex} presented \textit{MWELex}, a multilingual lexical of Croatian, Slovene, and Serbian multi-word expressions (MWE) that were extracted from parsed corpora. The \textit{srMWELex} lexicon v0.5 was automatically built during the short-term scientific mission inside the PARSEME COST action. It contains multi-word expression candidates extracted with the DepMWEx tool from the \textit{srWaC} v1.0 web corpus. It consists of 22 290 entries and 3 273 369 multi-word units. The freely available  morphological lexicon srLex is introduced in \citep{ljubevsic2016new}. It is consisting of 105 359 lexemes and 5 327 361 (token, lemma, MSD) triples. 

 \cite{miletic2017building} described the creation of a morphosyntactic e-dictionary for the Serbian language. It is derived from the Wiktionary edition for Serbo-Croatian, a manually POS-tagged corpus and specialized proposition list. This lexicon contains 1 226 638 million wordforms for 117 445 lemmas, corresponding to a total of 3 066 214 unique triples (wordform, lemma, MSD - morpho-syntactic description), and it is aimed for POS (part of speech) tagging and parsing tasks.

The DELAS-TOP and DELAS-PERS are dictionaries that respectively list geographic names and Serbian personal names \citep{krstev2008resources, pavlovic2004towards, grass2002description}. The dictionary of geographic names
DELA-TOP covers geographic concepts at the
level of a high-school atlas (approximately 20.000
toponyms, oronyms, and hydronyms with their
corresponding derivatives). The dictionary of personal names has been created from the list of the names of 1.7 million inhabitants of Belgrade as established in 1993. Based on this list, two dictionaries were
constructed: DELA-FName for the first names, and
DELA-LName for the last names \citep{vitas2003overview}.

The dictionary of librarianship and information sciences contains terminology \citep{kovavcevic2004bibliotekarski} used in the theory and practice of librarianship,  information sciences, and related fields in Serbian, English, and German.
The online version of the dictionary currently contains:
40,000 definitions (approximately 14,000 in Serbian);
900 definitions or annotations of terms that are part of library standards; 2,300 acronyms of international and national organizations and institutions; 190 addresses of relevant websites\footnote{http://rbi.nb.rs/srlat/dict.html}.

The electronic geological dictionary (GeolISSTerm) is a specially prepared taxonomy of basic geological concepts and terms, and it is used for IT needs as an elementary resource in the formation of domains in the Geological Information System of Serbia (GeolISS)\citep{stankovic2011development}.

 \citep{vujicic2014enriching} extended the SrpMD by 636 entries of simple words and  612 entries of MWE (multi-word expressions) from the culinary domain.

\cite{grljevic2016sentiment} provided several dictionaries for sentiment analysis in the field of education in her doctoral dissertation (sentiment words, domain-specific phrases, negation keywords, and stop words that are identified from the corpus).
Negation signals, negative quantifiers, and particle intensifiers were added to the sentiment lexicon \citep{ ljajic2019improving}. Similarly, for sentiment analysis, a domain-oriented stop words collection was created \citep{mladenovic2016}. In a separate chapter on sentiment analysis below, sentiment word lexicons and other lexicons used in sentiment analysis are described more. 

\cite{avdic2020automated} created medical dictionaries for Serbian: names of diagnoses (7942 entries), diagnosis code (14194 entries), Latin names of the diagnosis (3794 entries), therapies (2232 drugs and 1317 ampoules, 2255 different terms), symptoms for the diagnosis of measles B05 (95 entries), specialties (41 entries), abbreviations from the medical domain. Non-medical dictionaries created in the same research are a set of negation symbols in the medical domain, places, and names.

\cite{ostrogonac2020python} created a domain vocabulary of jobs in Serbian. It has two versions, one of 40 thousand, one of  80 thousand words, and 30 thousand lemmas, and they are included in Python library \textit{nlpheart}.

The Serbian stop word dictionary (SSW dictionary) contains 1241 different stop words for the Serbian language. It was created based on the grammar of the Serbian as well as by comparing with available sets of stop words for the Serbian language and a set of stop words for the Croatian language. SSW dictionary for the Serbian language contains words in different forms of their appearance. A word type label accompanies each word. The SSW dictionary is available as a CSV file - SSWdictionary.csv. The file contains two columns: word and label. The label describes the type of words: auxiliary verbs (V), pronouns (PRON), adverbs (ADV), prepositions (PREP), conjunctions (CONJ), exclamations (EXCL), particles (PART) and abbreviations (ABBR)\citep{marovac2021creating}.

The SrHurtLex \citep{stankovic2020multi} is a lexicon created for the detection of abusive words in Serbian. It is created using the lexical database Leximirka, the system of Serbian morphologic dictionaries SrpMD, and  The Dictionary of Serbian Language (DS) \citep{vujanic2007}, where the multi-word expressions labeled in dictionaries as augmentative, pejorative, derogatory, vulgar, etc. were collected.
 
\subsection{Ontologies}

The term "ontology" originates from philosophy and it represents science about existing concepts (types of things) and their relations \citep{vujivcic2016ekstrakcija}. In computer science, ontology is a structure that describes concepts, their relations, and existing constraints. Their purpose is the automatic sharing and reuse of knowledge between humans and computer, and between computers. Both parts which are included in sharing process have to have a certain level of understanding of the exchangeable information. 

The hierarchy of classes is called taxonomy. Commonly, ontology describes terms and relationships between them for a particular domain.

The semantic network which describes proper names and their relations is developed during Prolex project \citep{krstev2007note}. It consists of 2000 proper names, mainly names of states and their capital cities.

The RudOnto is a terminological resource developed at the Faculty of Mining and Geology in Belgrade, and it is the reference resource for mining terminology in Serbian. It is managed by a terminological information system, and 
intended to produce the derived terminological resources in subfields
of mining engineering, such as planning and management of exploitation, mine
safety or mining equipment management \citep{stankovic2011development}.

\cite{tomavsevic2018razvoj} developed a mining domain ontology \textit{RuDokOnto} for the purpose of collecting, describing, and systematization of mining project documentation throughout the phases of the mining project's life cycle in a way that links other related ontologies.

RetFig is a linguistic domain, descriptive, formal
ontology for rhetorical figures in Serbian and describes 98 figures \citep{mladenovic2013ontology}.

\subsection{Word networks}

In traditional dictionaries, lexical concepts are alphabetically ordered and there is a definition for all possible meanings for each of them. In WordNet, all words expressing a concept are grouped together in a set of synonyms (synset - synonymous set).\textit{ Serbian WordNet (SerWN)} \citep{krstev2004using, koeva2008morpho} is the lexical-semantic net for Serbian. Its development started within the project BalkanNet \citep{mladenovic2020two}, and when it finished in 2004, it had 8000 synsets. After that, the development of WordNet continued, especially in biological, biomedical, psycho-linguistic, and gastronomical domains, etc. Its structure is basically the same as PWN (Princeton Word Net \citep{miller2007wordnet}), and it is organized using nodes (synsets) and relations between them. Every word in synset is represented as an array of characters or literal, followed by the meaning of concrete literal in concrete synset. As a word can have multiple meanings, it can be part of multiple synsets. 

According to \cite{koeva2008morpho}, \textit{SerWN} consists of 13612 synsets, 23139 literals, 18210 relations, 314 derived, and 83 derivatives. 

 \cite{krstev2014approximate} developed an ontology for the culinary domain in Serbian, and the Serbian Wordnet is enhanced with the synsets from this domain. This ontology is used for the determination of similarity between recipes and query expansion.

As a lexical resource, \textit{SerWN} has been applied in multi-member lexical unit research  \citep{krstev2010automatic, mladenovic2014developing}, text classification \citep{pavlovic2010ontology}, the search of multilingual digital databases \citep{ranka2012tool}, recognizing rhetorical figures \citep{mitrovic2017ontological, mladenovic2013ontology, mitrovic2014electronic}, analyzing feelings expressed in the text \citep{mitrovic2015adding} and others.

Vujicic-Stankovic in  created an ontology for the culinary domain, and expanded SrWN by 1,404 synsets from the culinary domain so it contains a total of 1,797 such synsets \citep{vujicic2014enriching, vujivcic2016ekstrakcija}.

 Universal Dependencies (UD) project\footnote{\url{https://universaldependencies.org/introduction.html}} aims to develop cross-linguistically consistent treebank annotation for many languages, to provide a universal inventory of categories and guidelines to facilitate consistent annotation of similar constructions across languages while allowing language-specific extensions when necessary. As a part of this project, Serbian treebank is created, based on SETimes corpus \citep{samardvzic2017universal}.

\section{Lexical and syntactic analysis methods}

\subsection{Transliteration and diacritic restoration}\label{sec2}

The tool for the automatic performing diacritic restoration of text which is potentially missing diacritics
(e.g. transform "kuca" (dog) into "kuća" (house), if it is necessary) is described by \cite{ljubevsic2016corpus}. The accuracy of the tool is 99.5\% on standard and 99.2\% on nonstandard language.

Transliteration in Serbian is accommodated because each sound is a character. Characters map almost directly from Cyrillic to Latin, with exception of a few letters, that map from a single Cyrillic character to two Latin characters (e.g. њ -> nj, љ -> lj, or ђ -> dj). Systems for transliteration between Serbian Cyrillic and Latin alphabets exist since the 1950s \citep{matthews1952latinisation,aurousseau1953transliteration,gerych1965transliteration}. 
Among newer tools for solving this problem is the Python package \textit{nlpheart} \citep{ostrogonac2020python}, which has a possibility of conversion between the Cyrillic and Latin alphabet.

\subsection{Tokenization and stemming}

Sentence tokenization is the process of dividing the text into consisting sentences. Word tokenization's aim is to divide sentences into simple units, tokens, which are usually words, numbers, and punctuation marks. There are a number of multi-language tokenizers which have the ability to tokenize Serbian texts. The majority of these tools are available as Python modules, like Cutter\citep{graen2018cutter}, Spacy\footnote{\url{https://spacy.io/api/tokenizer}}, 
CLASSLA and Reldi\footnote{\url{https://github.com/clarinsi/reldi-tokeniser}} tokenizers. Cutter tokenizer has a variant for online tokenization\footnote{\url{https://pub.cl.uzh.ch/projects/sparcling/cutter/current/}}. CLASSLA tokenizer is adapted Stanford NLP Python Library with improvements for specific languages - Fork of Stanza for Processing Slovenian, Croatian, Serbian, Macedonian and Bulgarian)\footnote{\url{https://pypi.org/project/classla/}}. Turanjin tokenizer for Serbian is available as a PHP library\footnote{\url{https://github.com/turanjanin/serbian-language-tools}}. There is no precise information or comparison of the tokenization accuracy on Serbian documents.

\cite{ostrogonac2020python} present Python package \textit{nlpheart} for text processing of Serbian that includes transliteration, tokenization, normalization, and automatic preparing for the application of machine learning models. 

Stemming is a process of removing finishing letters of words, as derivation suffixes of words. The remaining part is a reduced form of the word called a stem. The stem differs from a dictionary form of the word (lemma).
The first tool for stemming (in further text, stemmer) for Serbian is described by \cite{kevselj2008suffix}, and it is rule-based (1000 rules) and  its accuracy is 79\%. Based on this stemmer,  \cite{milovsevic2012stemmer} created a new stemmer reducing the number of rules (180 rules) with an accuracy of 90\%. Another solution can be found in literature, and it is created by S. Petković et al.\footnote{Stefan Petković and Dragan Ivanović, Stemmer for Serbian language, 2019. \url{https://snowballstem.org/algorithms/serbian/stemmer.html} (accessed Apr 26, 2022)} and it is based on Stemmer for Croatian (precision of 0.986 and recall of 0.961 (F1 0.973) for Croatian)\footnote {Ljubešić, Nikola. Pandžić, Ivan. Stemmer for Croatian,  \url{http://nlp.ffzg.hr/resources/tools/stemmer-for-croatian/}}. There is no information about the stemming accuracy of this tool. Batanović et al. reimplemented the optimal and the greedy stemmers of \cite{kevselj2008suffix}, improved the greedy algorithm proposed by \cite{milovsevic2012stemmer}, and reimplemented a stemmer for Croatian by Ljubešić \& Pandžić, which is a refinement of the algorithm presented by \cite{ljubevsic2007retrieving}, as a WEKA package \citep{holmes1994weka} – SCStemmers in \citep{batanovic2016reliable}.

The stem is not a dictionary word form, it is the most common part of words with the same semantic meaning. So, in some normalization methods, n-gram analysis is used as a stemmer alternative. This means that a word could be normalized to a single sub-string of its letters whose size is n (tri-gram, tetra-gram etc.). The reason is that the n-gram analysis approach is language-independent, which means that it doesn't need any rules, lexicons, or corpora. \cite{marovac2012n} used n-gram analysis in the normalization of Serbian text. 

\subsection{Lemmatization and Part-of-speech tagging}

Lemmatization is a process that aims to determine the base morphological form of the word (lemma), which corresponds to a headword in a dictionary. This step in text mining is especially important for languages with rich inflectional morphology, such as Serbian. A given word can have multiple possible lemmas, and it depends on the context, so some lemmatizers use information obtained by POS or MSD tagging to achieve better accuracy.

There are a number of lemmatization approaches: rule-based, simple statistical-based methods, and machine learning-based methods \citep{akhmetov2020highly}. 

LemmaGen \citep{jurvsic2010lemmagen} is a learning algorithm for the automatic generation
of lemmatization rules in the form of a refined RDR (Ripple Down Rules) tree structure. It is compared with CST \citep{dalianis2006hand} and RDR \citep{plisson2008ripple} lemmatization algorithms and its lemmatization accuracy on Serbian  corpora are given in \ref{table3}. 

BTagger\footnote{https://github.com/agesmundo/BTagger} \citep{gesmundo2012lemmatising}  is a bidirectional tagger-lemmatizer tool that implements a lemmatization-as-tagging paradigm. Models are trained on the Serbian G.O.1984 corpus, reaching overall accuracies of 97.72\% for lemmatization and 86.65\% for MSD tagging.

\cite{agic2013lemmatization} tested hidden Markov model trigram taggers HunPos, lemmatization capable PurePos, TreeTagger, support vector machine tagger SVMTool, CST data-driven rule-based lemmatizer and BTagger on Serbian corpora and results are given in Table \ref{table3}.

\begin{table}[h!]
\tiny
\begin{center}
\begin{minipage}{300pt}
\caption{Tools for normalization and POS tagging}
\begin{tabular}
{@{}p{0.53\linewidth}p{0.12\linewidth}p{0.15\linewidth}p{0.22\linewidth}ll@{}}
 \hline
 Tool label & Application & Corpus & Accuracy \\ [0.5ex]
 \hline
   KeseljStemmer \citep{kevselj2008suffix} & stemmer & unknown & 79.0 \%  \\
MilosevicStemmer \citep{milovsevic2012stemmer}& stemmer 
 & Politika & 90.0\%   \\
\hline
 LemmaGen \citep{jurvsic2010lemmagen} & lemmatizer & Multext-East & 
 up to 86.1\%+-0.61\%  \\ 
CST \citep{jurvsic2010lemmagen, dalianis2006hand} & lemmatizer &  Multext-East & 64.0 \%+-0.82\%   \\
 RDR \citep{jurvsic2010lemmagen, plisson2008ripple} & lemmatizer &  Multext-East & 63.8\%+-0.80\%   \\
 BTagger \citep{gesmundo2012lemmatising} & lemmatizer & G.O.1984 & 97.73\%  \\
 PurePOS \citep{agic2013lemmatization} & lemmatizer & SETimes & 86.63\%  \\
  \hline
  TnT tagger \citep{gesmundo2012lemmatising} & POS tagger &  G.O.1984 & 85.47\%  \\
  BTagger \citep{gesmundo2012lemmatising} & POS tagger & G.O.1984 & 86.65\%  \\
HunPOS \citep{agic2013lemmatization} & POS tagger & SETimes &  95.47\% \\
 CRF \citep{ljubevsic2016new}  & POS Tagger & 500k & 97.86\%  \\
 \hline
 HunPOS \citep{agic2013lemmatization}  & MSD tagger & SETimes & 87.11\% (+lex 84.81\%) \\
PurePOS \citep{agic2013lemmatization} & MSD tagger &  SETimes & 74.4\% \\
 SVMTool \citep{agic2013lemmatization} & MSD tagger & SETimes & 84.99\%  \\
 CRF \citep{ljubevsic2016new} & MSD tagger & 500k & 92.33\% \\
 Reldi-tagger \citep{ljubevsic2019does} & MSD tagger &  CLARIN.SI & 92.03\%  \\
StanfordNLP \citep{ljubevsic2019does} & MSD tagger & CLARIN.SI & 95.23\%\\[1ex] 
 \hline
\end{tabular}
\label{table3}
\end{minipage}
\end{center}
\end{table}

POS (part of speech) tagging is an NLP processing task where words in the text are annotated with corresponding grammatical categories (parts of speech: verb, noun, adjective, pronoun, etc.). POS tagging with more precise information about grammatical categories is MSD tagging (morphosyntactic tagging - tagging with morphosyntactic descriptions). 

Finite state automata used in the lexical and syntactic analysis, considering morpho-syntactic labels were described in \citep{krstev1997jedan}.

\cite{sevcujskiautomatska} used HMM for morphosyntactic tagging on Alfanum and G.O.1984 corpora. The accuracy of annotation largely depends on the type of text and that some texts are more suitable for automatic annotation than others. For the AlphaNum corpus, an error of 18.44\% was obtained, and for "1984" as much as 26.97\%.

\cite{popovic2008evaluacija, popovie2010taggers} evaluated five taggers (Tree Tagger, SVMTool, Brill – Rule Based Tagger, Trigrams’n’Tags and MXPOST) on three corpora (\textit{“Helsinške sveske br. 15, nacionalne manjine i pravo”}, Serbian Radio diffusion Law and materials from UNDP workshops, G.O.1984).  TnT has shown the best performance, while Tree Tagger and SVMTool taggers have shown better performance in special cases.

The POS tagger for Serbian and Croatian based on CRF
(conditional random fields) is described in \citep{ljubevsic2016new}. It is trained on a manually annotated corpus of Croatian in combination with  hrLex/srLex lexicons for each language. The set of morpho-syntactic labels used in the corpus is created according to instructions of the revised MULTEXT-East V5 set of labels for Croatian and Serbian. The accuracy of POS tagging for Serbian is 92.33\% for MSD tagging and 97.86\ for POS tagging.

The tools for tokenization, stemming, lemmatization, and POS and MSD tagging and their accuracy on Serbian corpora are shown In Table \ref{table3}.

\section{Classification}

Text classification is a process of categorizing text data into predefined groups or categories based on its content. Text classification is often performed using supervised machine learning techniques.

\cite{graovac2014prilog} proposed two methods for classifying text based on their content. 
The first method is based on the representation of a document as a profile containing a fixed number of n-grams of bytes that appear in the document, and a dissimilarity measure used to determine the class to which the document belongs. This method is language-independent and does not require any pre-processing of the text or prior knowledge of the content of the text or the language in which the text is written. The second method refers to the use of the information contained in the Serbian wordnet and the Serbian electronic dictionary. 

Petrović  proposes utilizing models and neural networks as a potential remedy to meet the demand for machine prediction of links or references within the text of newly enacted laws and other regulations \citep{petrovic2019domain, petrovic2020analiza}. Training and validation of neural networks (RNN - Recurrent neural networks, CNN - convolutional neural networks, and HAN - hierarchical attention network model) are performed on a labeled data set, which is made by assigning to each segment of the text of the law (each article of the law) a corresponding label on the existence, or non-existence of a link or reference in that segment of the text. After that, the training procedure is based on a large set of data, which includes a collection of 1120 texts of laws, segmented into a total of 59 167 individual articles of law. 

For all methods, the number of training parameters is reduced by over 99\%.

\subsection{Similarity}

\cite{marovac2013similarity} proposed a method for similarity search of documents in Serbian. The searching query is represented as a word vector, as well as documents for search hing. The grouping of the documents is done using the k-Means clustering algorithm, and keywords are extracted using TF and IDF features, and n-grams. The similarity values between query and documents are calculated using cosine measure, Jaccard's coefficient, or Euclidian distance. \cite{furlan2013semantic} proposed a new algorithm, called LInSTSS, which, when determining the semantic similarity of two short texts, also takes into account the specifics of the words these texts contain. The evaluation was carried out on a corpus of paraphrases for the Serbian language created in the same research. One solution of similarity search in e-government is described by \cite{nikolic2016modelovanje} using the tool “Apache Lucene". \cite{petrovic2019influence} demonstrated how different preparation methods influence the calculation of text similarity.

\cite{batanovic2020metodologija} presented the process of handling semantic tasks using statistical modeling and machine learning. The STS.news.sr is a corpus of news created and used for the task of semantic similarity where the similarity of news is annotated by score. Implementation is given in the library STSFineGrain (Java), available on GitHub. For semantic similarity, the combination of word alignment and the average of word vectors was used. The srWaC corpus (Web corpus of the Serbian language) is used for creating the word vectors. An evaluation of the effects of 3 different stemming techniques on text similarity for Serbian has been performed. Additionally, a new technique for calculating similarity was proposed called Part-of-Speech and Term Frequency weighted Short-Text Semantic Similarity.

\subsection{Sentiment analysis}

Sentiment analysis is the process of analyzing and deriving people's opinions, thoughts, and impressions regarding various topics, products, and services expressed in a part of the text. Sentiment analysis can be investigated on several levels: document level, sentence level, phrase level, and aspect level \citep{wankhade2022survey}. For sentiment analysis, specific lexical resources are necessary, such as a dictionary of sentiment words, tools for processing negation, stylistic figures, and so on.
One of the first tools for sentiment analysis at the sentence level for the Serbian language was given by \cite{milovsevic2012mavsinska}. A binary classification of negative and positive sentiment was performed using the Naive Bayes(NB) algorithm.
 A steamer \citep{milovsevic2012stemmer} that was designed for this purpose was used as part of the preprocessing. Stop words were eliminated, and negation processing was done by prefixing the word that follows the negation signal (words like no, none) with 'NE\_'. The sentiment analyzer was created as a web tool and made available to the public\footnote{\url{https://inspiratron.org/SerbianSentiment.php}}.
 
Maximum entropy (ME), support vector machine (SVM), and NB machine learning methods were used to analyze tweet sentiment  \citep{jolic2015}. Procedures are offered to minimize the noise in these messages to increase accuracy. They achieved the best accuracy with the ME method of 80.5\% using unigrams; however, when applying unigrams and bigrams, negation and phrases were also considered, increasing accuracy to 82.7\%. 

 \cite{mladenovic2016hybrid} chose a hybrid approach that uses a dictionary of sentiments extended by morphological forms using a morphological dictionary SrbMD 
and synonyms using Serbian WordNet 
to reduce the disadvantages of using stemmers in morphologically rich languages. A sentiment dictionary was created \citep{mladenovic2016}, containing 1053 expressions (and 10704 inflectional forms) classified into 24 emotion categories, and augmented with synonyms and phrases. SentiWordNet has been integrated with Serbian WordNet to provide sentiment tags to the synsets from Serbian WordNet. A total of 4044 synsets were marked. An additional sentiment dictionary with 971 inflectional forms was created using these synsets. Using the TF-IDF approach, 577 (1428 inflectional forms) of the most frequent words from the 122 million-word corpus of the contemporary Serbian language SrpKor2013 were used to construct the list of stop words. A domain-oriented collection of stop words with 1372 inflectional forms were generated using the TF approach. The method was trained on a news set (TrN, Table \ref{tabsent}) with two topics: "bad news" and "good news," which are automatically categorized and balanced by sentiment. Two sets were used for testing: a set of news (TsN, Table \ref{tabsent}) collected from a source other than TrN (this set is not balanced), and a set of movie reviews (TsMR, Table \ref{tabsent}) collected from a website and tagged with the sentiment, based on the grades that were attached to them (this set is not balanced). These sources were used to develop the Serbian document-level sentiment analysis framework (SAFOS), which applies the maximum entropy approach with the features: of unigrams, bigrams, and trigrams. They used hold-out test sets and 10-fold cross-validation (CV) to evaluate the SAFOS system. The combination of unigram and bigram features reduced by "sentiment feature" mapping produced the best classification accuracy scores for both hold-out tests (accuracy 78.3\% for TsMR set and 79.2\% for the TsN set). Because it was trained and tested on data from the same domain, it performed better in a 10-fold CV with a 95.6\% accuracy rate.

 \cite{grljevic2016sentiment} presented a sentiment analysis of content from social networks to improve the business of higher education institutions. Sentiment analysis is performed at two levels of granularity: at the document level and the sentence level. On the set of online reviews of professors and lectures (ORPL, Table \ref{tabsent}) both a rule-based strategy (based on a vocabulary that was manually built for the requirements of this domain and is available), as well as the approach based on machine learning algorithms (NB, SVM, and k-Nearest Neighbor KNN) were applied. In sentiment classification using machine learning algorithms, the SVM algorithm gives the best performance, with 84.94\% accuracy at the review level, and 80.13\% accuracy at the sentence level. The classification of the sentiment was done using the sentiment lexicon, by introducing separate dictionaries for 1266 positive and 1521 negative sentiment words, intensifiers (95), neutralizers, negation (31), domain-
specific phrases (41), stop words (179), and other words that change the sentiment of the next word in the sentence. The classification accuracy at the level of reviews is 80.71\% and at the level of sentences, it is 73.70\%.

The first balanced and topically uniform sentiment analysis dataset in Serbian (SerbMR, Table \ref{tabsent}) was generated by  \cite{ batanovic2016reliable} and is available online in versions with two sentiment polarity classes (positive and negative; 1682 documents) and three polarity classes (positive, neutral, and negative; 2523 documents). The sentiment labels, in this dataset, were obtained automatically by converting the numerical ratings attached to each review by its author. This dataset was examined to identify the best machine-learning features and simple text-processing options for sentiment classification. By combining the obtained optimal attributes with NBSVM (combination of polynomial Naive Bayes classifier and support vector method classifier), they achieved an accuracy of up to 85.55\% for two and up to 62.69\% for three classes. By comparing different methods for morphological normalization, it was concluded that the use of stemmer is better than lemmatization in the case of sentiment analysis. The stemmer of Ljubešić and Pandžić gave the best accuracy results on the dataset SerbMR, 86.11\% for two and up to 63.02\% for three classes \citep{batanovic2017sentiment}.

According to the studies cited above, identifying the presence of negation is insufficient to ascertain sentiment. The collection of film reviews in \citep{batanovic2016reliable} is subjected to the traditional method of processing negation, which involves changing the polarity of words that follow a negative signal. For three classes, marking two words after the negation led to the most significant improvement in sentiment analysis accuracy (0.94\%), while for two classes, marking only the first word after the negation gave the best improvement in accuracy (0.66\%). The processing rules of semantic negation, which improved the classification of short informal texts by sentiment, are described in \cite {ljajic2019improving}. These rules were tested on a set of tweets with topic public personalities that were manually marked with the sentiment (TWPP, Table \ref{tabsent}).
The machine learning method that uses additional attributes based on the proposed negation processing rules improves sentiment analysis accuracy on a set of tweets for three classes by up to 1.45\% and for two classes by up to 0.82\%. When this method is applied to a set of tweets containing negation, the improvement in sentiment analysis accuracy increases
 by up to 2.65\% for three classes and
up to 1.65\% for two classes. For this study's aims, dictionaries of negation signals (25), negative quantifiers (56), and intensifiers, as well as a sentiment dictionary of 5632 sentiment words (reduced to the morphological foundation of 4058 negative and 1574 positive words), were constructed. The impact of various morphological normalizations on sentiment analysis was examined on this set of tweets, and it was discovered that the use of stemmer \citep{milovsevic2012stemmer} takes precedence over normalization using the morphological dictionary SrbMD (accuracy 85.27\%) and that reducing words to 4-grams produces good results with little resource  usage \citep{ljajic2019comparison}.

Aspect-based sentiment analysis deals with the identification of sentiments (negative, neutral, positive) and the determination of aspects (target sentiments) in a sentence.  \cite{nikolic2020aspect} proposed an aspect-based sentiment analysis of student opinion surveys in the Serbian language. Two sets of data were used for sentiment analysis, which was done at the finest level of granularity of the text - the level of the sentence segment (phrase and sentence).

A collection of official student surveys (OSS, Table \ref{tabsent}) makes up the first dataset, while the second dataset set of online reviews of professors and lecturers (OSPL, Table \ref{tabsent}) previously created for the paper \citep{grljevic2016sentiment}.
The OSS and OSPL corpora were automatically annotated  for the sentiment (negative, neutral, positive), then manually annotated for aspects (ranging from lower-level features, such as lectures, helpfulness, materials, and organization, to higher-level aspects, such as professor, course, and other).
For aspect classification, a cascade classifier (a collection of SVM binary classifiers trained to distinguish between two distinct aspects) was employed. The quality of the aspect analysis was influenced by the corpus, as seen by the F-measures of 0.89 for the OSS corpus and 0.78 for the OSPL corpus, respectively.

\begin{table}[h!]
\tiny
\begin{center}
\begin{minipage}{300pt}
\caption{Corpora for sentiment analysis}\label{tabsent}%
\begin{tabular}{@{}p{0.2\linewidth}p{0.18\linewidth}p{0.15\linewidth}lp{0.2\linewidth}@{}}
\toprule
Corpus label & Text type & Number of items & Annotation\footnotemark[1]  & Reference  \\
TrN   & News   & 2000 & S; AA  & \text{\citep{ mladenovic2016hybrid}} \\
TsN                 & News                               & 779  & S; AA        & \text{\citep{ mladenovic2016hybrid}} \\
TsMR                 & Movie reviews                      & 2237 & S; AA        & \text{\citep{mladenovic2016hybrid}} \\

ORPL & Education Reviews & 3863 & S, A; AA + MA & \text{\citep{grljevic2016sentiment}}\\
OSS  & Education Reviews & 2472 & S,A; AA +MA                                              & \text{\citep{nikolic2020aspect}}    \\

SerbMR               & News                               & 2523 & S;AA                                                  & \text{\citep{ batanovic2016reliable}} \\
SentiComments.SR     & Short texts                        & 3490 & S; MA &
\text{\citep{ batanovic2020metodologija}}             \\
ParlaSent-BCS   v1.0 & Sentences of parliamentary debates & 2600 & S; MA                                                 & \text{\citep{ mochtak2022parlasent}}  \\
TWPP                 & Tweets                             & 7664 & S; MA                                                 & \text{\citep{ ljajic2019improving}}    \\
TWVA                 & Tweets                             & 8817 & S,R; MA                                            & \text{\citep{ ljajic2022uncovering}}\\ 
MRSA & Music Reviews    & 1830 & S; AA                                            & \text{\citep{ draskovic2022development}}\\ 
SMSSA & SMS messages    & 6171 & S; MA                                            & \text{\citep{vsandrih2019sms}}\\ 
TW15 & Tweets    & 1643735  & S; MA                                            & \text{\citep{11356/1054}}\\ 

\botrule
\end{tabular}
\footnotetext[1]{Annotation target: S - sentiment, A - aspect, R - relevance;
Annotation type: AA - automated annotation MA - manually annotated }

\end{minipage}
\end{center}
\end{table}

Sentiment analysis includes specific subtasks such as polarity detection, subjectivity detection, sarcasm detection, etc. An annotation approach with six sentiment labels was created to satisfy the requirements of processing particular tasks and enabling multiple interpretations of sentiment   \citep{batanovic2020versatile}. SentiComments.SR (Table \ref{tabsent}), a corpus of short texts in the Serbian language, has been manually annotated using this multi-level annotation scheme. It contains 3490 short movie comments (length up to 50 tokens) \citep{batanovic2020metodologija}. On this corpus, the outcomes of applying linear classifiers using bag-of-words and/or bag-of-embedding features were evaluated under the influence of different morphological normalizations and negation processing techniques. The combination of bag-of-words and bag-of-embeddings attributes resulted in significant improvements in classification for all sentiment analysis subtasks (F - measure: polarity 0.783, subjectivity  0.885 four-class sentiment analysis 0.655, six-class sentiment analysis   0.586). Due to the insufficient number of sarcastic texts in the corpus, the results of sarcasm detection are not representative.

Sentiment lexicon Senti-Pol-sr \citep{stankovic2022sentiment} was created based on three existing lexicons (NRC, AFFIN, and Bing) and was manually corrected.
The dictionary contains 6454 different tokens. Its initial version is available. 

The lexicon was utilized to conduct sentiment analysis on a well-balanced dataset extracted from SrpELTeC, which consisted of 1089 sentences that were manually labeled, with each sentiment category containing 363 instances of positive, neutral, and negative sentiments. This approach achieved the best accuracy of 87.8\% on SrpELTeC 2 classes and 71.9\% on SrpELTeC 3 classes using MNB with the Bag-of-Words approach combined with our sentiment lexicon features.
The results of trained models using LR, NB, decision tree, random forest, SVN, and k-NN methods gave the best accuracy of 87.8\% for LR. It has also been shown that training on a dataset of labeled movie reviews (SerbMR) indicates that it cannot be successfully used for sentence sentiment analysis in old novels.
 \cite{draskovic2022development} developed a machine-learning model for sentiment analysis using three different data sets.
 The first set (MRSA, Table \ref{tabsent})  was created for this research by collecting music reviews from 13 portals, which made sure that the set was balanced. The second data set is the already mentioned set of movie reviews, while the third set is music album reviews—MARD.
MARD was originally composed in English and then translated into Serbian using the Google Translate API. Standard classification models (NB, LR, and SVM) and hybrid models (combining a linear model with NB) were applied to these datasets. The hybrid model NB-LR gave average good results (58\% for three classes and 79\% for two classes). It is shown that a set of film and music reviews can be used together to improve the quality model. Extending the model with reviews translated from English does not improve performance, due to the different vocabulary and review writing styles, as well as the quality of the translated text.
Emoticon influence, informal speech, lexical, and other language features about the mood in the set of SMS messages (SMSSA, Table \ref{tabsent}) are presented by \cite{vsandrih2019sms}. They selected 621 features and divided them into three main categories:
lexical (based on signs and words), syntactic (emoticons, abbreviations), and stylistic.
Using linear SVM classification, an accuracy of 92.3\% was obtained. Sentence-based sentiment classification as well as emotion recognition is suggested to improve the classification of SMS messages.
 \cite{mozetivc11smailovi} found that the quality of the classification model depends much more on the quality and size of the training data than on the type of trained model by analyzing 1.6 million manually tagged tweets in 15 different European languages, of which 73,783 tweets are in Serbian (TW15, Table \ref{tabsent}). Based on the performed experiments, it was shown that there is no statistically significant difference between the performance of the top classification models (five of these models are based on SVM, and for reference, the NB classifier was applied).

Transfer learning is one of the advanced techniques in AI, which allows a pre-trained model to transfer its knowledge to a new model.  Transfer learning is frequently used in sentiment analysis to classify sentiments, and it can produce successful results, particularly in the absence of large labeled data sets. 

Batanovic presented the results of applying neural language models based on transformer architectures to sentiment analysis subtasks of short texts from the SentiComments.SR corpus \citep{batanovic2020versatile}.Three transformer-based models were used: Multilingual BERT \citep{devlin2018bert}, Multilingual distilBERT \citep{sanh2019distilbert}, and XLM MLM  \citep{conneau2019cross}.
Fine-tuning multilingual transformer-based models gain the same or better performance than linear models for all sentiment analysis subtasks. For each subtask, XLM MLM produced the best F-measure results: 
 0.793 for polarity, 0.887 for subjectivity, 0.686 for four-class sentiment analysis, and 0.627 for six-class sentiment analysis.

Based on a sample of parliamentary discussions,  \cite{mochtak2022parlasent} demonstrated that using transformer models produces outcomes that are noticeably superior to those obtained using a simpler architecture. 
The dataset consists of sentences of average length from the corpus of parliamentary proceedings in the region of the former Yugoslavia - Bosnia and Herzegovina, Croatia, and Serbia. A set of 2600 sentences (ParlaSent-BCS   v1.0, Table \ref{tabsent}), including 876 with only positive, 876 with only negative, and 866 without sentiment words, were chosen for the dataset using the Croatian gold standard sentiment lexicon \citep{glavavs2012semi} (translated to Serbian with a rule-based Croatian-Serbian translator \citep{klubivcka2016collaborative}). This dataset contains 1059 sentences from the Serbian parliament. The dataset is manually annotated using the multiple-level annotation schema described by \cite{batanovic2020versatile}, and it is available online. A sentiment analysis approach was applied at the sentence level.
   The results of classification four of the transformer models were compared:
FastText \citep{bojanowski2017enriching} with pre-trained CLARIN.SI word embeddings \citep{ljubevsic2018word},
  XLM-R \citep{conneau2019unsupervised},
  CroSloEngual BERT \citep{ulvcar2020finest}, and BERTi\'{c} \citep{ljubevsic2021berti}.
BERTi\'{c} gave the best results compared to the others (model macro F1 0.7941 ± 0.0101). Compared to Bosnian and Croatian, the Serbian language proved to be the most difficult to predict.
Using BERTi\'{c} for sentiment analysis,  \cite{ljajic2022uncovering}  expanded the annotated dataset that was used for the topic analysis of tweets containing negative sentiment towards the COVID-19 vaccination. A collection of 8817 vaccination-related tweets in the Serbian language (TWVA, Table \ref{tabsent}) were manually labeled as relevant or irrelevant regarding the COVID-19 vaccination sentiment. Relevant tweets were manually marked with sentiment labels: positive, negative, or neutral. On this data set, BERTi\'{c} correctly categorized tweets as relevant or irrelevant with a 94.7\% accuracy rate and correctly classified relevant tweets as negative, positive, or neutral with an 85.7\% accuracy rate. The annotated set was expanded by this classifier, and from the original manually annotated 1770 tweets with negative sentiment, another 1516 tweets with negative sentiment were automatically marked, forming the data set used for the topic analysis. The topic analysis was carried out using the latent Dirichlet allocation (LDA) and nonnegative matrix factorization (NMF) methods.
Topics that are potential reasons for vaccine skepticism are highlighted by topic analysis: worries about adverse reactions, efficacy, inadequate testing, mistrust of authorities, and conspiracy theories.

\section{Named entity recognition}

Named-entity recognition (NER) is a task that seeks to locate and classify named entities mentioned in unstructured text into categories such as personal names, organizations, locations, medical codes, time expressions, quantities, monetary values, percentages, etc. Named entity recognition (NER) as an NLP task is fairly old, gaining popularity with Message Understanding Conferences in the mid-1990s \citep{sekine2004named}, However, NER for Serbian has not been addressed substantially until the 2010s. 

\cite{vitas2008resources} developed a system that uses morphological and lexical analysis in combination with dictionaries (Serbian and transcribed English first names, and geographical locations) for recognition of people's names and geographical entities. The system is using e-dictionaries and transducer-based rules or grammars for disambiguation of proper names and geopolitical entities \citep{krstev2007note,vitas2008resources}.

\cite{ljubevsic2013combining} proposed a first system based on machine learning and conditional random fields for the recognition of names, organizations, and locations for Croatian and Slovene, which are closely related to Serbian. They have used a set of annotated web and news corpora (SETimes, Vjesnik, and corpora for both Slovene and Croatian developed as a student project \citep{filipic2012strojno}) to train their method. For features, they used linguistic features and distributional similarity features calculated from large unannotated monolingual corpora. Their experiments showed that distributional features improve the F1 score by 7-8 points, while morphological features can improve by additional 3-4 points. However, as the size of the dataset increases, the morphosyntactic and distributional features lose their importance for NER. They have made resources used for building this NER system publicly available. 

Another approach, based on the previous application of rules encoded in transducers and thesauri \citep{krstev1997jedan} was enhanced for recognition of personal names and geopolitical names \citep{krstev2014system}. Dictionaries are used for matching tokens and phrases, while recursive transition networks (grammar graphs) from Unitex \citep{paumier2002unitex} are used to resolve ambiguities (e.g. taking into account grammatical rules such as case-number-gender agreement). They reported that the system prefers precision over recall, with a precision of 0.96 and a recall of 0.88. 

For the purpose of comparing NER approaches in multi-lingual aligned texts (bitexts), a system called NERosetta was developed \citep{krstev2013nerosetta,krstev2013nerosetta2}.  To illustrate the system, 7 bitexts involving 5 languages (French, English, Greek, Serbian, Croatian) and 5 different NER systems were used (1 for Serbian \citep{krstev2014system}, 1 for Croatian \citep{ljubevsic2013combining}, 1 for English (Stanford NER) and 2 for French). The entities that were evaluated were Person, Organization, and Location, with some of the NER systems providing annotations for time, date, money, percent, and others. The demo application is available on the web \footnote{\url{http://www.korpus.matf.bg.ac.rs/nerosetta/}}. 

A dictionary approach with the addition of transducer-based grammars \citep{krstev2014system} was used to create a gold standard data set based on news articles annotated with personal names. This data set was then used to train machine learning-based approaches, namely Stanford NER and SpaCy \citep{vsandrih2019development}. Their evaluation indicated that the rule-based approach performed the best (based on the F1-score), while Stanford NER had the best recall. 

\cite{tanasijevic2019toward} developed a system for labeling cultural heritage documents with metadata. In order to do this, she developed a system that recognizes entities, such as years and person names, as well as topics of the tagged documents. 

The transformer-based model was also introduced for several tasks in Serbian, Croatian, and Slovene, including NER \citep{ljubevsic2021berti}. The model was pre-trained on web-crawled texts in Serbian, Bosnian, Croatian, and Slovene, consisting of 8 billion tokens, and then fine-tuned for NER on several openly available datasets, such as SETimes.SR \citep{batanovic2018setimes}, corpora of news articles, or ReLDI-sr \citep{ljubevsic2017serbian}, corpora of annotated tweets. For reference, authors compared this model with CroSloEngual BERT \citep{ulvcar2020finest} and multilingual BERT \citep{devlin2018bert}, where language-specific BERT-based models significantly outperformed multi-lingual BERT. 

Apart from a general domain for Serbian, a decent amount of work has been done for medical named entity recognition. One of the previously described systems for a general domain was adapted for medical de-identification of clinical texts \citep{jacimovic2014automatic,jacimovic2015rule}. The system recognized persons, dates, geographic locations, organizations, and numbers using vocabularies and transducer grammar rules. The authors reported an overall F1 score of 0.94. On the other hand, \citep{puflovic2016supervised} created a model based on character and word n-grams. The dataset they used was obtained from a neurological clinic and their system was designed to recognize names of diseases, names of medications, abbreviations, and numbers that represent dosage, dates or times, and medical treatment success. They have manually checked 100 documents, reporting accuracy ranging from 64\% to 90\%. A mathematical model for medical term recognition was proposed by \citep{avdic2020automated}. They have proposed three methods. The first method was based on the dictionary matching of terms. The second method uses a formula for labeling words contained in the training set, where confidence is
calculated by the number of instances in which a word is labeled with a certain label in the training set divided by the total count of that word in the training set. The third method is an extension of the second method, where several rules are added to terms with errors and abbreviations, so they can be tagged well. Labeled entities included medical terms (symptoms, symptom descriptions, diagnoses, biochemical analyses, Latin words, anatomic names of organs, therapies, and other medical terms) and non-medical terms (numbers, negation symbols, and other words). The best-performing model was the third one, with an F1 score of 0.937, while the highest F1 score for medical terms was 0.896. The methods based on deep neural networks and multilingual language models were proposed by \cite{kaplar2022evaluation}. They used a manually annotated corpora from the Clinic for Nephrology at the University Clinical Center of Serbia (203 discharge summaries annotated by 2 computer science Ph.D. students adapting the 2012 i2b2 temporal relation challenge annotation schema). They have created models based on conditional random fields (CRF), multilingual transformers (BERT Multilingual and XLM RoBERTa), long short-term memory (LSTM) recurrent neural networks, and their ensembles. CRF method had hand-crafted features that are commonly used in literature (word, word stem, shape of the word, previous 3 words, next 3 words, etc.). For the LSTM model, the authors used  gensim’s word2vec model before feeding the embeddings to the LSTM network, followed by the CRF token classifier. The study showed that the highest precision was achieved with the CRF-based model, while the highest recall had a multi-lingual transformer model. The best F1 score had an LSTM-CRF-based model. The best performance was achieved by creating an ensemble of the models with majority voting (F1 score of 0.892). 

\section{Language models}

A language model determines word probability in a sequence. In order to create a language model, many approaches were proposed, ranging from simply calculating word appearance in a larger text corpus to adding more lexical and syntactic features to learning word probabilities. Early language models were purely statistical, while since 2014, we have seen a proliferation of neural language models - language models based on neural networks. 

Language models are prerequisites for many natural language tasks. Therefore, many works in classification \citep{graovac2014prilog}, sentiment analysis \citep{milovsevic2012mavsinska,jolic2015,grljevic2016sentiment,batanovic2017sentiment,ljajic2019improving} or named entity recognition \citep{vsandrih2019development}, used traditional n-gram language models, at times enriched with lexical, morphological or syntactic features. These systems were previously described in this review.

 \cite{ostrogonac2018modeli} in his PhD thesis does a review and comparison of language models for Serbian up to 2018. In this work, he proposes the first neural language model for Serbian, based on recurrent neural networks trained on a corpus of morphologically annotated text in Serbian.  Also, he creates a hybrid model that uses parts-of-speech and lemmas, and matches sequences of words to either n-grams in corpus, or to partially lemmatized sequences. These models are compared with more traditional n-gram models for correcting semantic and grammatical errors in the  text. The error is detected by setting a threshold for a difference in log likelihood between a language model with morphological features and one without it. While setting thresholds may be challenging, it showed the potential use cases for specifically trained neural language models for Serbian. 

There has been a significant effort done by international researchers to create multilingual neural language models. Some of these models included also Serbian, such as FastText \citep{bojanowski2017enriching}, multilingual BERT \citep{devlin2018bert}, XLM-R \citep{conneau2019unsupervised}, and XLM MLM \citep{conneau2019cross}. Batanovi\'{c} in his Ph.D. thesis \citep{batanovic2020metodologija}, compared a number of n-gram language models for the tasks of sentiment analysis and text similarity. He further compared these language models and methods with fine-tuned multi-lingual transformer-based models (multilingual BERT base \citep{devlin2018bert}, DistilBERT Multilingual \citep{sanh2019distilbert}, and XLM MLM \citep{conneau2019cross}), showing transformer models in all cases outperforming all n-gram based models (including ones containing a large amount of morphological, lexical, and syntactic features). 

The first, and, at the time of writing of this paper, the only transformer-based language model specifically trained for Serbian, Croatian, Bosnian, and Montenegrin is BERTi\'{c} \citep{ljubevsic2021berti}. BERTi\'{c} is trained using the ELECTRA approach \citep{clark2020electra} for training transformer models. This approach involves training a smaller generator model and the main discriminator model with the task to discriminate whether a specific word is an original word from the text or a word generated by the generator model. The model is trained on a corpus of 8 billion tokens crawled from the web in Serbian, Croatian, Bosnian, and Montenegrin. While there was previously a BERT-based model for Croatian and Slovenian, called CroSloBERT \citep{ulvcar2020finest}, BERTi\'{c} outperformed it on almost all tasks (morphological annotation, NER, social media geolocation prediction, commonsense causal reasoning task). This is mainly because of the bigger corpus, and computational efficiency of the ELECTRA approach that was used.


\section{Conculsion and future directions}

Research on natural language processing for the Serbian language has a long tradition, going back to the second half of the 1990s. During this time, many approaches for lexical, morphological, syntactic, and semantic processing of text were explored. In the past decade, the number of researchers and research published on natural language processing for Serbian significantly increased. Several Universities and research institutes in Serbia established natural language research groups. 

The Serbian language is a highly inflected language and therefore many challenges in natural language processing are specific to Serbian, such as the most efficient way for tokenization, handling the inflections in various tasks, handling negations, etc. While some work has been done on these challenges, they are still open research questions. Basic lexical and morphological tasks, such as transliteration, diacritic restoration, tokenization, stemming, lemmatization, and part-of-speech tagging are quite well-researched, with many approaches presented, evaluated, and compared. Some of the classification tasks, such as sentiment analysis were, as well, extensively researched. Sentiment analysis seems to gain a lot of interest after 2012. Named entity recognition has also been researched for several named entities, such as proper and personal names, and locations. Also, few approaches have been proposed for biomedical NER. 

On the other hand, some methods and tasks were still not adequately addressed for Serbian. Many classification tasks, except sentiment analysis, have not been explored and language resources for them are missing. As it was previously said, methods for only a basic set of named entities have been proposed. Domain-specific classification and named entity recognition methods are still missing. 

Methods in the semantic web, ontology, and semantic networks have not much proliferated in the area of Serbian NLP, as only a few papers are touching on this subject. Most significant research in network space has been done in developing Serbian WordNet, but this is a rather morphological and lexical network, then something that can be considered a semantic network. Language resources for many of the tasks are still missing. 

While the language-specific BERT-based model has been trained, there is only a single initiative to create this kind of language model. Also, resources such as sentence embedding or document embedding methods have not been yet developed. These methods would also contribute significantly to the creation of methods for summarization, question answering, language-specific semantic search, or machine translation. 

At the moment, there is a proliferation of large language models, such as GPT-3 \citep{brown2020language}, Lambda \citep{thoppilan2022lamda} or ChatGPT \citep{ouyang2022training}. While these models are multilingual and can generate text in Serbian, there has not yet been much research on prompt engineering or fine-tuning these language models for Serbian. 
\backmatter

\section{Acknowledgements}
This paper is partially supported by the Ministry of Education, Science, and Technological Development of the Republic of Serbia, Projects No. III44007.

\bibliographystyle{plainnat}

\bibliography{sn-bibliography}


\end{document}